\pdfoutput=1

\documentclass[11pt]{article}

\usepackage[]{ACL2023}

\usepackage{times}
\usepackage{latexsym}
\usepackage{amsmath}
\usepackage{ulem}
\usepackage{booktabs}
\usepackage{multirow}
\usepackage{enumitem}
\usepackage{algorithm}
\usepackage{algorithmic}
\usepackage{amssymb}
\usepackage{bbding}
\usepackage[inkscapelatex=false]{svg}
\usepackage[T1]{fontenc}

\usepackage[utf8]{inputenc}

\usepackage{microtype}

\usepackage{inconsolata}

\title{CoT-MAE v2: Contextual Masked Auto-Encoder with Multi-view Modeling for Passage Retrieval}

\author{Xing Wu\textsuperscript{\rm 1,2,3 }, Guangyuan Ma\textsuperscript{\rm 1,2 }\thanks{The first two authors contribute equally.} , Peng Wang\textsuperscript{\rm 1,2 } \\ \textbf{Meng Lin}\textsuperscript{\rm 1,2 }, \textbf{Zijia Lin}\textsuperscript{\rm 3}, \textbf{Fuzheng Zhang}\textsuperscript{\rm 3} and \textbf{Songlin Hu}\textsuperscript{\rm 1,2}\thanks{Corresponding author.} \\
 \textsuperscript{\rm 1} Institute of Information Engineering, Chinese Academy of Sciences \\
    \textsuperscript{\rm 2} School of Cyber Security, University of Chinese Academy of Sciences \\
    \textsuperscript{\rm 3} Kuaishou Technology \\
\{wuxing, maguangyuan, wangpeng, linmeng, husonglin\}@iie.ac.cn \\
linzijia07@tsinghua.org.cn, zhangfuzheng@kuaishou.com
}

\begin{document}
\maketitle

\begin{abstract}
Growing techniques have been emerging to improve the performance of passage retrieval. As an effective representation bottleneck pre-training technique, the contextual masked auto-encoder utilizes contextual embedding to assist in the reconstruction of passages. However, it only uses a single auto-encoding pre-task for dense representation pre-training. This study brings multi-view modeling to the contextual masked auto-encoder. 
Firstly, multi-view representation utilizes both dense and sparse vectors as multi-view representations, aiming to capture sentence semantics from different aspects. Moreover, multi-view decoding paradigm utilizes both auto-encoding and auto-regressive decoders in representation bottleneck pre-training, aiming to provide both reconstructive and generative signals for better contextual representation pre-training.
We refer to this multi-view pre-training method as CoT-MAE v2. Through extensive experiments, we show that CoT-MAE v2 is effective and robust on large-scale passage retrieval benchmarks and out-of-domain zero-shot benchmarks. 
\end{abstract}

\section{Introduction}
Passage retrieval involves searching a large corpus for passages that are relevant to a particular query, and is essential for various applications such as web search \cite{fan2021pre, guo2022semantic, lin2021pretrained}, question answering \cite{karpukhin2020dense, lee2020learning, zhu2021adaptive}, and dialogue systems \cite{gao2022neural, yu2021few}.
Sparse retrieval methods, such as BM25, are practical and dominant approaches. 
However, retrieval methods based on pre-trained language models (PLMs) \cite{devlin2018bert, liu2019roberta} have gained popularity in recent years. PLMs are utilized to improve (1) sparse retrieval models that rely on lexical match; (2) dense retrieval models that model the semantic interaction between queries and passages in latent semantic space.

\begin{figure*}
\centering
\includegraphics[width=16cm]{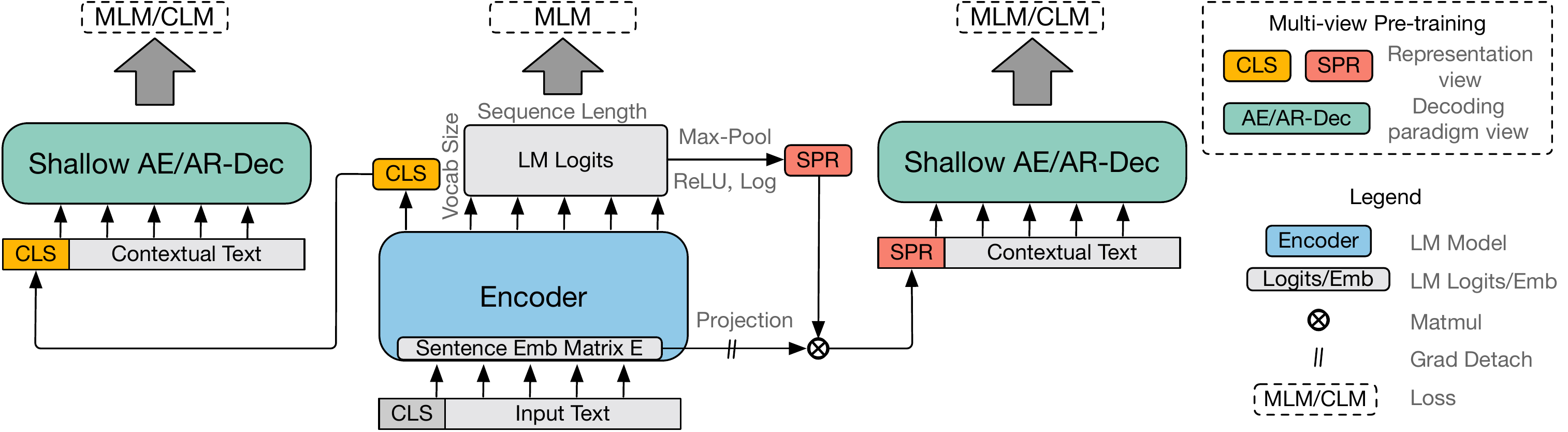}
\caption{
Pre-training designs of CoT-MAE v2. CoT-MAE v2 utilizes both dense (CLS) and sparse (SPR) vectors as multi-view representations. As a multi-view decoding paradigm, Auto-Encoding Decoder (AE-Dec) and Auto-Regressive Decoder (AR-Dec) are integrated into contextual masked auto-encoder pre-training to provide both MLM reconstruction signals and CLM generative signals for representation pre-training. 
}
\label{cotmae-v2-model}
\end{figure*}

To improve sparse retrieval, PLM mitigates vocabulary mismatch by projecting each term in the query and passage to a vocabulary-sized weight vector. Each dimension in the weight vector represents the weight of a term in the PLM vocabulary, and the correlation between query and passage depends on lexical matching \cite{formal2021splade, formal2021spladev2, shen2022lexmae}. PLM-based sparse vectors model the features of passage on every lexical term (denoted by PLM logits output), and thus naturally capture the semantic at the \textbf{\textit{lexicon level}}. SPLADE \cite{formal2021splade, formal2021spladev2} uses learnable PLM, e.g., BERT, to produce sparse vectors and fine-tune the retriever with contrastive learning loss. PLM-based SPLADE outperforms traditional sparse retrieval like BM25.
LexMAE \cite{shen2022lexmae} further expands on SPLADE by proposing to pre-train a lexical-bottlenecked masked auto-encoder to learn importance-aware sparse representations. 
It learns the lexicon importance distribution in an unsupervised way, through a continuous bag-of-words representation bottleneck. 

PLM-based dense retrieval typically employs a siamese or dual-encoder architecture to convert queries and documents into a low-dimensional vector space \cite{hofstatter2021efficiently, humeau2019poly, xiong2020approximate, zhan2021optimizing, zhan2020repbert}. Relevance between queries and documents is calculated using cosine similarity or dot products. This low-dimension vector is called the dense vector and is trained to capture the sentence semantics at the \textbf{\textit{passage level}}. Recent efforts aim to improve dense retrieval performance by adding auxiliary tasks to pre-training, like contextual-supervised learning \cite{gao2021unsupervised, wu2022contextual}.
Contextual-supervised learning aims to improve text representation by considering the context of surrounding passages. A recent method, CoT-MAE \cite{wu2022contextual}, expands on this by incorporating contextual masked auto-encoding, leading to better dense retrieval performances.
CoT-MAE involves selecting two passages from a document as contextual passages, aiming to learn a better representation with the assistance of contextual information. A shallow encoder, typically with one or two layers, is used to reconstruct the masked passages with the help of context passage embedding.

The efficiency of PLM-based sparse retrieval \cite{bai2020sparterm, formal2021splade, formal2021spladev2, shen2022lexmae} and dense retrieval \cite{wu2022contextual, wang2022simlm, gao2021condenser} demonstrates that both sparse and dense representations effectively capture the text meaning. 
On the basis of contextual masked auto-encoder (CoT-MAE) tailored for dense retrieval, a natural thought is to incorporate the sparse representation pre-training into CoT-MAE, i.e., exploring multi-view representations in contextual masked auto-encoder pre-training and fine-tuning.
Theoretically, sparse representation focuses on the lexicon while dense representation focuses on sentence semantics, offering different perspectives on the text. These two representations are probably compatible and complement each other.

Aside from representation variations, there is also a distinction in the structure of the decoder in the previous methods. Some \cite{wu2022contextual, shen2022lexmae} choose to use auto-encoding decoders for pre-training tasks, while others \cite{lu2021less, liu2022retromae} prefer to use auto-regressive decoders.
Despite being distinct, these decoders are capable of utilizing the text representations from the encoder for pre-training, leading us to investigate pre-training with multiple decoders. The two types of decoders have different inductive biases and vary in their usage of text representation. We believe that, if a text representation is good enough, it should be able to tolerate the usage variation, enabling any type of decoder to accomplish the pre-training task effectively.



Based on the above two assumptions, in this paper, we expand on CoT-MAE to bring multi-view modeling into contextual masked auto-encoder pre-training.\\
$\bullet$ \textbf{Multi-view representation} proposes using dense and sparse representation in pre-training, fine-tuning, and inferencing, aiming at jointly capturing the sentence semantics from both passage and lexicon levels. By further incorporating contextual representation bottleneck pre-training, multi-view representation can achieve strong retrieval performance.\\
$\bullet$ \textbf{Multi-view decoding paradigm} proposes to provide unbiased self-supervision signals to contextual representation pre-training. Both auto-encoding and auto-regressive decoders are used in contextual masked auto-encoders to assist the pre-training of contextual embedding through Masked Language Modeling (MLM) reconstructive signals and Casual Language Modeling (CLM) generative signals. The joint usage of both decoding signals contributes to a steady improvement in retrieval performance.


The improved method that incorporates the fusion of these views is referred to as \textbf{CoT-MAE v2}, as shown in Figure \ref{cotmae-v2-model}.
To verify the effectiveness of our proposed multi-view contextual masked auto-encoder pre-training, we conduct experiments on large-scale web search benchmarks: MS-MARCO Passage Ranking \cite{nguyen2016ms}, TREC Deep Learning (DL) Track 2019 \cite{craswell2020overview} and Track 2020 \cite{Craswell2020OverviewOT}. 
We also evaluate on the BEIR \cite{thakur2021beir} benchmark with a large set of out-of-domain datasets.
Experimental results show that CoT-MAE v2 has considerable gains over competing baselines and achieves the new state-of-the-art performance on these benchmark datasets.

Our contributions can be summarized as follows: \\
 $\bullet$ We propose to incorporate the multi-view pre-training technique to contextual masked auto-encoder pre-training. \\
 $\bullet$ Multi-view representations and multi-view decoding paradigms provide better semantics capturing and rich self-supervision signals to representation learning. \\
 $\bullet$ Experiments show that CoT-MAE v2 brings considerable performance gains over competing baselines in passage retrieval.



\section{Related Works}
Retrieval methods based on pre-trained language models (PLMs) \cite{devlin2018bert, liu2019roberta} have recently gained popularity.
\subsection{PLMs for Dense Retrieval}
PLMs improve dense retrieval in both the pre-training stage and fine-tuning stage. 
\paragraph{\textbf{Pre-training stage}} Attempts in the pre-training stage are split into two categories. The first category focuses on passage prediction tasks for passage retrieval \cite{chang2020pre, gao2021unsupervised, ma2022pre, wu2022contextual, zhou2022master}.
For instance, \cite{chang2020pre} pre-trains with inverse cloze task (ICT), body first selection (BFS), and wiki link prediction (WLP) tasks.
\cite{gao2021unsupervised} pre-trains with contrastive span prediction based on document proximity.
\cite{ma2022pre} extends contrastive span prediction to multiple granularities.
\cite{gao2021unsupervised} pre-trains with contrastive span prediction based on document proximity.
\cite{wu2022contextual} pre-trains with generative span prediction with mixed related span selection strategies.
The other category focuses on enhancing the encoder using auxiliary auto-encoding tasks \cite{lu2021less, gao2021condenser, liu2022retromae, wu2022contextual}. For instance, \cite{lu2021less, gao2021condenser} aims to boost text representation by auto-encoding with a weak decoder with limited capacity and attention flexibility. \cite{liu2022retromae, wu2022contextual, zhou2022master} improve text representation through asymmetric masking ratios applied to the encoder and decoder, then reconstructing the aggressively masked text with the help of its embedding or context embedding.
The most related method is \cite{wu2022contextual}, our work is on the basis of \cite{wu2022contextual} and incorporates the multi-view pre-training technique to \cite{wu2022contextual}.
\paragraph{\textbf{Fine-tuning stage}} In the fine-tuning stage, various methods have been used to improve performance, including mining hard negatives \cite{xiong2020approximate, zhan2021optimizing}, late interaction \cite{khattab2020colbert}, knowledge distillation  \cite{lin2021batch, santhanam2021colbertv2}, query clustering \cite{hofstatter2021efficiently}, data augmentation \cite{qu2020rocketqa}, and joint retriever-reranker optimization \cite{ren2021rocketqav2, zhang2022hlatr, zhang2021adversarial}. For instance, \cite{xiong2020approximate} creates hard negatives through ANN updates; \cite{zhan2021optimizing} improves negatives with a fine-tuned retriever; \cite{khattab2020colbert} models similarity with MaxSim on encoder hidden states; \cite{lin2021batch} distill knowledge from a MaxSim operator or strong re-ranker \cite{santhanam2021colbertv2}; \cite{hofstatter2021efficiently} introduces topic-aware query and margin sampling for efficiency; \cite{qu2020rocketqa} combines cross-batch negatives, denoised negatives, and data augmentation; \cite{ren2021rocketqav2} adaptively improves retriever and re-ranker with listwise distillation; \cite{zhang2022hlatr} incorporates retrieval-reranking features with HLATR; \cite{zhang2021adversarial} optimizes retriever-ranker through a minimax adversarial objective. The most related methods are \cite{xiong2020approximate, zhan2021optimizing, santhanam2021colbertv2}, our fine-tuning process is on the basis of these three methods.

\subsection{PLMs for Sparse Retrieval}
Attempts utilizing PLMs to improve sparse retrieval can be roughly divided into two categories.
\paragraph{\textbf{Term weighting}} The term weighting process aims to assign importance to terms based on contextualized token representations \cite{dai2020context, dai2021context, gao2021coil, lin2021few}. For instance, \cite{dai2020context} uses BERT token representations to determine the context-specific importance of terms in passages. \cite{dai2021context} extends this by splitting documents into passages and aggregating term weights. It uses titles, web links, and pseudo-relevance feedback to generate weak supervision signals for learning term weights. \cite{gao2021coil} computes the relevance score between query and text encodings by taking the dot product of overlapping terms. \cite{lin2021few} propose a framework to unify these approaches, which reduces \cite{gao2021coil}'s weight vector to one dimension, and can retain its effectiveness while increasing efficiency.

\paragraph{\textbf{Term expansion}} Term expansion improves vocabulary matching by using PLMs to expand queries or documents \cite{nogueira2019doc2query, formal2021splade, formal2021spladev2, shen2022lexmae}. The final representation of a text or query is obtained by combining token weight vectors, effectively expanding it to include non-occurring terms. A sparsity regularization is applied to achieve a sparse representation for efficient inverted index usage. For instance, \cite{nogueira2019doc2query} predicts relevant queries for a document. \cite{formal2021splade} and \cite{formal2021spladev2} project terms in queries and texts to a weight vector, estimated by logits of masked language models. \cite{shen2022lexmae} further expands on \cite{formal2021spladev2} by proposing to pre-train a lexical-bottlenecked masked auto-encoder to learn importance-aware sparse representations.

\section{Approach}
This section will introduce detailed pre-training, fine-tuning, and inference designs, including multi-view representations with both dense and sparse vectors, and the multi-view decoding paradigm for pre-training.

\subsection{Multi-view Representations} \label{Multi-view-representations}
Transformer-based pre-trained language model (PLM), e.g., BERT encoder (denote as $Enc$), is utilized to produce vectors for representing a whole text. Multi-view representations with dense and sparse vectors model the input sequences from different aspects. Dense vectors, e.g., $[CLS]$, naturally represent the sequence- or passage-level semantics. In contrast, sparse vectors produced by the PLM in the vocabulary dimension denote the logit frequency, and thus model the sentence from the lexicon level. Formally, Given a sequence of $N$ tokenized subwords as input text.

\begin{equation}
\mathbb{T} = \{CLS, t_1, ..., t_N, SEP\} \label{eq_1}
\end{equation}

The input text is forwarded through the $L$-layer Transformer-based BERT encoder. For the Transformer Layer $l \in \{1, ..., L\}$, the output hidden states are

\begin{equation}
\mathbf{H}^{l} = \{\mathbf{h}_0^{l}, \mathbf{h}_1^{l}, ..., \mathbf{h}_N^{l}\}
\end{equation}

\paragraph{\textbf{Dense representations}}
Following common practice, we usually take the first output hidden vector (CLS) of the last layer hidden state as the dense representation ($\mathbf{DEN}$).

\begin{equation}
\mathbf{DEN} = \mathbf{h}_0^{last}
\end{equation}

\paragraph{\textbf{Sparse representations}}
Sparse vector depicts BERT logits as frequencies. BERT is pre-trained as a masked language model by taking original masked-position logits as labels via Cross-Entropy (CE) loss. Thus the output of BERT in vocabulary size (e.g., BERT-base is 30522) naturally models the frequency of each logit for the input sequence. High frequency on one logit denotes a high occurrence frequency characterized by the BERT model, whose parameters are learnable. For sparse vectors, hidden states of the last layer are first projected to the vocabulary dimension logits with the transposed embedding matrix $\mathbf{E^\top}$.

\begin{equation}
\mathbf{w}_j^{i} = \mathbf{h}_i^{last}\mathbf{E}_j + b_j
\end{equation}

\noindent where $i \in \{1, ..., N\}$ and $N$ is the sequence length, $j \in \{0, ..., V-1\} $ and $V$ is the vocabulary size. $\mathbf{w}_j^{i}$ denotes individual logits in vocabulary dimension. Following SPLADE \cite{formal2021splade, formal2021spladev2}, max-pooling is performed on each logit along the sequence input axis to aggregate a max logit.

\begin{equation}
s_{j} = \max_{i \in \{1, ..., N\}} \log{(1 + ReLU(\mathbf{w}_j^{i}))}
\end{equation}

\noindent ReLU function will remove the negatives and keep positive numbers as logit frequencies. And the monotonically increasing log saturation function $\log{(1+x)}$ shrink down the high frequencies to ensure the sparsity. Finally, the sparse vector is denoted as follows.

\begin{equation}
\mathbf{SPR} = \{s_0, s_1, ..., s_j\}, j \in \{0, ..., V-1\}
\end{equation}

\paragraph{\textbf{Similarity Measurement}}
Sentence similarity measurement (denoted as $d(q, p)$) of queries (q) and passages (p) is computed via dot product or cosine similarity of corresponding representation vectors. Given a sentence-level dense vector $\mathbf{DEN}$ and a token-level sparse vector $\mathbf{SPR}$, here we denote the sum of two similarity measures as our multi-view representation similarity.

\begin{equation}
\label{distance measure of DEN and SPR}
d(q, p) = \mathbf{SPR}_q \cdot \mathbf{SPR}_p + \mathbf{SPR}_q \cdot \mathbf{SPR}_p
\end{equation}

Due to the extremely high storage cost of sparse vectors (e.g., BERT-base 30522 * 4 bytes for a single sparse vector), it's infeasible to store the whole vectors as continuous arrays directly. Previous works like SPLADE \cite{formal2021splade, formal2021spladev2} rounded the vectors as discrete integers and made virtual documents for sparse retrieval (e.g., BM25), which is improper to incorporate in the multi-view representation similarity measurement. Thanks to the sparsity, we apply the Top-${k}$ function that only preserves the $k$ highest frequencies in a sparse vector during fine-tuning and inferencing stages. Top-${k}$ keeps the storage to an affordable limit while preserving the most high-frequency logits in the vector space.

\subsection{Contextual Masked Auto-encoder with Multi-view Decoding Paradigm} \label{dec_paradigm}
Pre-training with contextual masked auto-encoding gives strong initialization to retriever models. Following Cot-MAE \cite{wu2022contextual}, we employ asymmetric encoder-decoder assisted with contextual embedding to give self-supervision signals to the representation vectors during pre-training. 

\paragraph{\textbf{Encoder}}
For the BERT encoder, given the input text $\mathbb{T}$ in Equation \ref{eq_1}, we replace a certain percentage of tokens in $\mathbb{T}$ with a specified mask token, e.g., $[MASK]$.
\begin{equation}
mask(\mathbb{T}) = \{CLS, t_1, MASK, t_3, ..., t_N, SEP\}
\end{equation}

\noindent We denote the index set of all $[MASK]$ positions as $\mathbb{M}$. The pre-training for BERT encoder (Enc) utilizes Cross-Entropy (CE) loss to optimize the following loss function, i.e., Masked Language Modeling (MLM) objective.

\begin{equation}
\mathcal{L}_{enc}=-\sum_{i \in \mathbb{M}} \log p(t_i|Enc(mask(\mathbb{T})))
\end{equation}


\paragraph{\textbf{Decoder}}
For the decoder side, we employ two different Transformers, Auto-Encoding Decoder (AE-Dec) and Auto-Regressive Decoder (AR-Dec), to provide multi-view pre-task signals for both dense and sparse vectors.
For pre-training with the dense vector, we directly take the concatenation of the dense vector with contextual sentence embedding and feed it into the AE-Dec and AR-Dec. As for pre-training with the sparse vector, we first project the $\mathbf{SPR}$ vector to hidden size dimension via production with grad-detached \footnote{This makes sure that the grad is only backpropagated through $\mathbf{SPR}$ vector, making steady parameters updates.} embedding matrix $\mathbf{E^\top}$. Then we take the concatenation of the sparse vector and feed it into the decoders. The inputs for decoders ($\mathbb{T}_{Dec}$) are denoted as two sequences used individually \footnote{For the purpose of clarity, we concatenate the dense (or sparse) vector and the tokens into a sequence in our subsequent formulas. In actual implementation, it is necessary to integrate the dense (or sparse) vector and the corresponding token embeddings.}.

\begin{equation}
\mathbb{T}_{Dec-DEN} = \{\mathbf{DEN}, t_1, ..., t_N, SEP\}
\end{equation}
\begin{equation}
\mathbb{T}_{Dec-SPR} = \{\mathbf{SPR} \cdot \mathbf{E}^\top, t_1, ..., t_N, SEP\}
\end{equation}

AE-Dec takes the Transformers Encoder layer as an Auto-Encoding decoder, which utilizes MLM as reconstruction signals. The loss for AE-Dec is formulated as

\begin{equation}
\mathcal{L}_{ae-dec}=-\sum_{i \in \mathbb{M}} \log p(t_i|Dec(mask(\mathbb{T}_{Dec})))
\end{equation}

AR-Dec takes the Transformers Decoder layer as an Auto-Regressive decoder, which utilizes Casual Language Model (CLM) as generative signals. The loss for AR-Dec is formulated as

\begin{equation}
\mathcal{L}_{ar-dec}=-\sum_{i \in \mathbb{T}} \log p(t_{i+1}|Dec(\mathbb{T}_{Dec}))
\end{equation}

Together with the MLM loss of the BERT encoder, both MLM reconstruction signals from AE-Dec and CLM 
generative signals from AR-Dec are used for multi-view effective pre-training.

\begin{equation}
\mathcal{L}_{total} = \mathcal{L}_{enc} + \mathcal{L}_{ae-dec} + \mathcal{L}_{ar-dec}
\end{equation}

\begin{figure*}
\centering
\includegraphics[width=16cm]{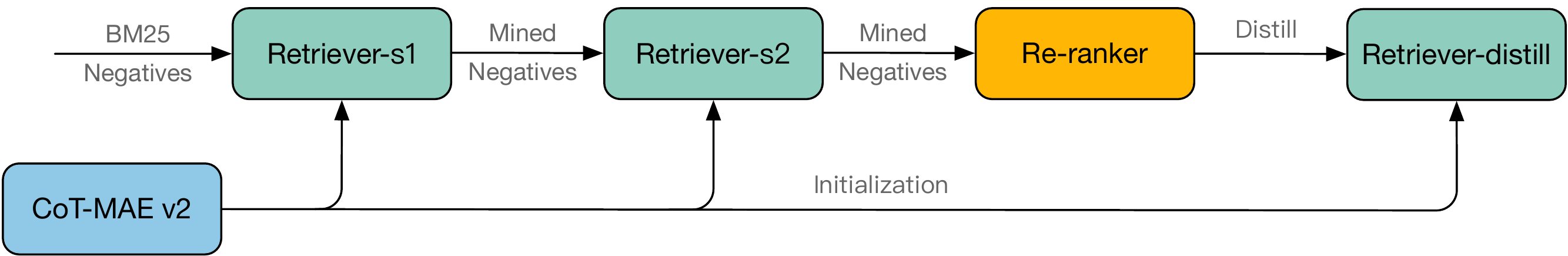}
\caption{
Illustration of our multi-stage fine-tuning pipeline. In stage one (s1), the retriever is trained with BM25 negatives. In stage two (s2), we use retriever-s1 to mine effective hard negatives for fine-tuning a retriever-s2. The scores tested on the retriever-s2 are used as non-distilled results. In stage three (re-ranker), a more powerful single-tower re-ranker is trained. In the last stage (distill), the re-ranker performs knowledge distillation to retriever-distill. 
}
\label{cotmaev2_finetune}
\end{figure*}

As argued in CoT-MAE, contextual embeddings on the decoder side effectively promote learning the BERT encoder's representations. Thus we use the contextual passages as the decoder input $\mathbb{T}_{Dec}$. For detailed contextual passage sampling methods, please refer to CoT-MAE \cite{wu2022contextual}.

\subsection{Fine-tuning and Inference}
\paragraph{\textbf{Fine-tuning}}
Fine-tuning is conducted on the downstream retrieval tasks to verify the effectiveness of pre-training. We use the multi-view representation similarity function $\mathbf{d(q, p)}$ as the distance measurement. Following DPR and SPLADE \cite{formal2021splade, formal2021spladev2}, we train both dense and sparse retrievers with a contrastive learning loss function.

\begin{equation}
\mathcal{L}_{\mathrm{ft}}=-\log \frac{\exp({d(q, p^{+})}) / \tau)}{\sum \exp(({d(q, p^{+})} + {d(q, p^{-})}) / \tau)}
\end{equation}

\noindent where $\tau$ is a temperature hyper-parameter fixed to 1. $\mathbf{p^{+}}$ is the human-labeled positive passages and $\mathbf{p^{-}}$ is the hard negative sets provided via BM25 or dense retrieval. 
Top-${k}$ sparse vectors are used in fine-tuning and inferencing stage. Following SPLADE \cite{formal2021splade, formal2021spladev2}, we also introduce a FLOPs regulation term to suppress the dominant logit frequencies in sparse retriever fine-tuning. Please refer to the SPLADE paper for FLOPs details. 

Following CoT-MAE \cite{wu2022contextual}, we trained retrievers on MS-MARCO Passage Ranking Task through multi-stage pipelines, as shown in Figure \ref{cotmaev2_finetune}. In stage one (s1), the retriever is trained with BM25 negatives provided by the MS-MARCO training set. In stage two (s2), we use retriever-s1 to mine effective hard negatives for fine-tuning the retriever-s2. The performance of retriever-s2 is our \textbf{non-distilled} result. In stage three (re-ranker), a more powerful single-tower re-ranker is trained. We directly use the CoT-MAE re-ranker 
 without re-training. In the final stage (distill), the abilities of the re-ranker are transferred to the retriever-distill through the process of knowledge distillation using KL loss.

\begin{equation}
\begin{aligned}
 & \mathcal{L}_{\mathrm{ft-distill}}  \\
  &=\sum_{q} KL({d(q, p | retriever)}, {d(q, p | reranker)})  
\end{aligned}
\end{equation}

\paragraph{\textbf{Inference}}
For the inferencing stage, we forward the queries and passages corpus through retrievers, to get the CLS dense vectors and Top-${k}$ sparse vectors. We use equation ({\ref{distance measure of DEN and SPR}}) to measure the multi-view query-passage similarities. We use Faiss for dense retrieval and PyTorch Sparse matrix multiplication for sparse retrieval.

Following BEIR \cite{thakur2021beir}, we train the retriever with MS-MARCO negatives for the out-of-domain evaluation on BEIR benchmarks.

\section{EXPERIMENTAL SETTINGS}
In this section, we introduce our experimental settings, including pre-training, fine-tuning, inference and baseline methods.
\subsection{Pre-training}
\paragraph{\textbf{Datasets}}
Following CoT-MAE \cite{wu2022contextual} and BERT \cite{devlin2018bert}, MS-MARCO documents and Wiki+BookCorpus are used as pre-training corpus. \\
$\bullet$ \textbf{MS-MARCO} documents contain 3.2 million documents. \\
$\bullet$ \textbf{Wiki+BookCorpus} contains 5.6 million documents. \\
NLTK sentence tokenizer was used to split each document and dynamic pad to target max length. Maxlen for MS-MARCO is 144 and Wiki+BookCorpus is 512 for alignment with MS-MARCO and BEIR benchmark passages. Contextual passages are selected as described in \cite{wu2022contextual} for contextual masked auto-encoder pre-training.

\paragraph{\textbf{Implementation}}
Following CoT-MAE \cite{wu2022contextual}, the encoder is initialized with a pre-trained 12-layer BERT-base model, while the decoders (including AE-Dec and AR-Dec) are initialized as a single layer from scratch. We pre-train the model using the AdamW optimizer for a maximum of 50k steps, with a learning rate of 4e-4, a batch size of 16k, and a linear schedule with a warmup ratio of 0.1. We use 16 Tesla A100 GPUs to train the model for 49 hours, and then discard the decoder, leaving only the encoder for fine-tuning. We set a widely used random seed as 42 for reproducibility. We also do statistical significance analysis to prove the robustness and effectiveness of our method. 

\subsection{Fine-tuning and Inference}
\paragraph{\textbf{Datasets and Evaluation}} 
We fine-tune the retrievers initialized from our pre-trained encoder on MS-MARCO passage datasets. Then we do the evaluation on MS-MARCO passage ranking \cite{nguyen2016ms}, TREC Deep Learning (DL) Track 2019 \cite{craswell2020overview} and 2020 \cite{Craswell2020OverviewOT} tasks. We also do out-of-domain evaluations on BEIR benchmarks.

MS-MARCO \cite{nguyen2016ms} is a benchmark dataset that contains 8.8 million passages and real user queries collected from Bing search. The training set contains 500k human annotated query-passage pairs, while the dev set contains 6,980 annotated queries. Since the test set is not publicly available, the dev set is used for evaluation. Following previous work\cite{wu2022contextual}, we evaluate the performances on MS-MARCO using MRR@10, Recall@50, and Recall@1K. 

TREC Deep Learning (DL) \cite{craswell2020overview, Craswell2020OverviewOT} tracks provide test sets with more elaborate annotations to evaluate the real capacity of ranking models. We evaluate the 2019 and 2020 test sets. The 2019 test set contains 200 annotated queries and the 2020 test set contains 200 annotated queries. Following previous work\cite{wu2022contextual}, we evaluate the performance on TREC with NDCG@10.

BEIR \cite{thakur2021beir} is a heterogeneous benchmark that contains 18 datasets on nine task domains. Following its fine-tuning setting, we fine-tune the retrievers with MS-Marco hard negatives\footnote{\url{https://sbert.net/datasets/msmarco-hard-negatives.jsonl.gz}}. We evaluate the performances on 14 publicly available datasets for out-of-domain BEIR evaluation.

\paragraph{\textbf{Implementation}}
Fine-tuning is trained on the MS-MARCO dataset and evaluated on the MS-MARCO dev set and TREC DL 2019/2020 test sets. Following \cite{wu2022contextual}, we trained on 8 Tesla A100 GPUs using the AdamW optimizer with a learning rate of 2e-5, a global batch size of 64, an epoch number of 3, and a weight decay of 0.01. The query length is 32, the passage length is 144, the number of negative passages is 15 and the negative depth is 200. The fine-tuning random seed is also set to 42 for reproducibility. For sparse retriever fine-tuning, the Top-${k}$ of sparse vectors is set to 768, and the FLOPs factor is set to 0.002.

Fine-tuning for BEIR evaluation is also trained on 8 A100 GPUs with the AdamW optimizer. Following \cite{thakur2021beir}, the learning rate is 1e-5, the global batch size is 1k, the epoch number is 10 and the number of negative passages is 1. Other hyperparameters are the same as the above settings.

\begin{table*}[t]
\centering
\small
\begin{tabular}{l|ccc|c|c}
\toprule  
& \multicolumn{3}{c|}{\textbf{MS-MARCO}} & \textbf{TREC DL 19} & \textbf{TREC DL 20}\\
\textbf{Model} & MRR@10 & R@50 & R@1k & nDCG@10 & nDCG@10 \\
\midrule
\midrule
BM25 $\dagger$ &  18.7  & 59.2  & 85.7  & 51.2 & 47.7 \\
SPLADE $\dagger$ \cite{formal2021splade}  & 32.2  & - & 95.5  & 66.5 & - \\
SEED \cite{lu2021less}  & 33.9  & - & 96.1  & - & - \\
ColBERT \cite{khattab2020colbert}  & 36.0  & 82.9  & 96.8  & - & -\\
RocketQA \cite{qu2020rocketqa}  & 37.0  & 85.5  & 97.9  & - & - \\
coCondenser \cite{gao2021unsupervised}  & 38.2  & 86.5  & 98.4  & 71.7 & 68.4 \\
SimLM \cite{wang2022simlm}  & 39.1 & - & 98.6 & - & - \\
RetroMAE \cite{liu2022retromae}  & 39.3  & - & 98.5  & - & - \\
CoT-MAE \cite{wu2022contextual}  & 39.4 & 87.0 & 98.7 & 70.9 & 70.4 \\
LexMAE $\dagger$ \cite{shen2022lexmae}  & 40.8  & - & 98.5  & - & - \\
\midrule
\textbf{CoT-MAE v2}  & \textbf{41.4} & \textbf{89.4} & \textbf{98.7} & \textbf{74.5} & \textbf{70.8} \\
\bottomrule
\end{tabular}
\caption{
Main results of \textit{non-distilled} retrievers on MS-MARCO passage ranking and TREC DL datasets. The best score is marked in \textbf{bold}. Models marked with $\dagger$ are sparse retrieval models.
}
\label{table_results_main}
\end{table*}

\begin{table*}[t]
\centering
\small
\begin{tabular}{l|ccc|c|c}
\toprule  
& \multicolumn{3}{c|}{\textbf{MS-MARCO}} & \textbf{TREC DL 19} & \textbf{TREC DL 20}\\
\textbf{Model} & MRR@10 & R@50 & R@1k & nDCG@10 & nDCG@10 \\
\midrule
\midrule
SPLADE v2 $\dagger$ \cite{formal2021spladev2}  &  36.8 & - & 97.9 & 72.9 & - \\
RocketQA v2 \cite{ren2021rocketqav2}  &  38.8 & 86.2 & 98.1 & - & - \\
AR2 \cite{zhang2021adversarial}  &  39.5 & 87.8 & 98.6 & - & - \\
ColBERT v2 \cite{santhanam2021colbertv2} &  39.7 & 86.8 & 98.4 & - & - \\
ERNIE-SEARCH \cite{lu2022ernie_search} &  40.1 & 87.7 & 98.2 & - & - \\
CoT-MAE \cite{wu2022contextual}  &  40.4 & 88.5 & 98.7 & - & - \\
SimLM \cite{wang2022simlm}  &  41.1 & 87.8 & 98.7 & 71.2 & 69.7 \\
MASTER \cite{zhou2022master} &  41.5 & 88.6 & 98.8 & 72.7 & 71.7 \\
LexMAE $\dagger$ \cite{shen2022lexmae}  &  42.6  & - & \textbf{98.8}  & 73.7 & \textbf{72.8} \\
\midrule
\textbf{CoT-MAE v2}  &  \textbf{43.1} & \textbf{90.2} & 98.7 & \textbf{75.6} & 72.5 \\
\bottomrule
\end{tabular}
\caption{
Main results of \textit{distilled} retrievers on MS-MARCO passage ranking and TREC DL datasets. The best score is marked in \textbf{bold}. Models marked with $\dagger$ are sparse retrieval models.
}
\label{table_results_main_distill}
\end{table*}

\subsection{Baselines}
Our baseline methods include the non-distilled retrieval method and distilled retrieval method, as shown in Table \ref{table_results_main}.
We compare the non-distilled retrievers with the latest state-of-the-art baseline methods, including sparse retrieval BM25 \cite{robertson2009probabilistic}, SPLADE \cite{formal2021splade}, dense retrievers SEED \cite{lu2021less}, ColBERT \cite{khattab2020colbert}, RocketQA \cite{qu2020rocketqa}, coCondenser \cite{gao2021unsupervised}, SimLM \cite{wang2022simlm}, RetroMAE \cite{liu2022retromae}, and CoT-MAE \cite{wu2022contextual}. \\
We compare the distilled retrievers with multiple strong distilled baselines, including sparse retriever SPLADE v2 \cite{formal2021spladev2}, dense retrievers RocketQA v2 \cite{ren2021rocketqav2}, AR2 \cite{zhang2021adversarial}, ColBERT v2 \cite{santhanam2021colbertv2}, SimLM \cite{wang2022simlm}, MASTER \cite{zhou2022master} and LexMAE \cite{shen2022lexmae}. \\

\section{EXPERIMENT RESULTS}
In this section, we analyze the experimental results to demonstrate the effectiveness of the proposed CoT-MAE v2 method.

\subsection{Main Results}
\paragraph{\textbf{Non-disilled Results}} Main results in Table \ref{table_results_main} show that non-distilled CoT-MAE v2 outperforms multiple latest strong baselines. Non-distilled retrievers are fine-tuned on BM25 negatives or retriever-mined negatives. For performances with non-distilled retrievers, CoT-MAE v2 outperforms the previous baseline method LexMAE by +0.6 points on MRR@10, outperforms CoT-MAE by 2.4 points on R@50 of the MS-MARCO passage ranking task, and outperforms CoT-MAE by +3.6 points on nDCG@10 of the TREC-DL 19 task. \\
\paragraph{\textbf{Disilled Results}} Main results in Table \ref{table_results_main_distill} show that distilled CoT-MAE v2 achieves state-of-the-art performances over the latest strong re-ranker distilled baselines. For instance, comparing the competing distilled retrievers, CoT-MAE v2 outperforms LexMAE by +0.5 points on MRR@10, outperforms MASTER by +1.6 points on R@50 of the MS-MARCO passage ranking task, outperforms LexMAE by +1.9 points on nDCG@10 of the TREC-DL 19 task. The distilled CoT-MAE v2 achieves 43.1 on MS-MARCO MRR@10, which even beats many power re-rankers and is approximate to its teacher CoT-MAE re-ranker (MRR@10 43.9). These performance gains clearly show the leading advantages to previous baseline methods. 


\begin{table*}[htbp]
\centering
\small

\begin{tabular}{l|c|c|c|c|c|c|c|c}
\toprule  
\textbf{Dataset} & BM25 & DPR & ANCE & ColBERT & Contriever & RetroMAE & MASTER  & \textbf{CoT-MAE v2} \\
 \midrule
TREC-COVID & 65.6  & 33.2  & 65.4  & 67.7  & 59.6  & 75.6 & 62.0 & \textbf{77.1}  \\ 
NFCorpus & 32.5  & 18.9  & 23.7  & 30.5  & 32.8  & 30.1 & 33.0 & \textbf{33.5}  \\ 
 \midrule
NQ & 32.9  & 47.4  & 44.6  & 52.4  & 49.8  & 49.0  & 51.6 & \textbf{53.9}  \\ 
HotpotQA & 60.3  & 39.1  & 45.6  & 59.3  & 63.8  & 63.8  & 58.9 & \textbf{67.2}  \\ 
FiQA-2018 & 23.6  & 11.2  & 29.5  & 31.7  & 32.9  & 30.1 & 32.8 & \textbf{33.1}  \\ 
 \midrule
ArguAna & 31.5  & 17.5  & 41.5  & 23.3  & 44.6  & 48.1 & 39.5 & \textbf{48.2}  \\ 
Touché-2020 & \textbf{36.7}  & 13.1  & 24.0  & 20.2  & 23.0  & 24.3 & 32.0 & 30.3  \\ 
 \midrule
CQADupStack & 29.9  & 15.3  & 29.6  & 35.0  & 34.5  & \textbf{38.2} & 32.7 & 32.2  \\ 
Quora & 78.9  & 24.8  & 85.2  & 85.4  & \textbf{86.5}  & 85.6 & 79.1  & 86.1  \\ 
 \midrule
DBPedia & 31.3  & 26.3  & 28.1  & 39.2  & 41.3  & 38.5  & 39.9 & \textbf{42.6}  \\ 
 \midrule
SCIDOCS & 15.8  & 7.7  & 12.2  & 14.5  & \textbf{16.5}  & 15.0  & 14.1 & \textbf{16.5}  \\ 
 \midrule
FEVER & 75.3  & 56.2  & 66.9  & 77.1  & 75.8  & 71.9  & 69.2 & \textbf{81.2}  \\ 
Climate-FEVER & 21.3  & 14.8  & 19.8  & 18.4  & 23.7  & 21.4  & 21.5 & \textbf{27.5}  \\ 
SciFact & 66.5  & 31.8  & 50.7  & 67.1  & 67.7  & 64.8  & 63.7 & \textbf{69.2}  \\ 
 \midrule
Average & 43.0  & 25.5  & 40.5  & 44.4  & 46.6  & 46.9  & 45.0 & \textbf{49.9} \\ 
\bottomrule
\end{tabular}
\caption{Out-of-domain evaluation on BEIR benchmark. The score that is better in comparison is marked in \textbf{bold}.}
\label{table_beir_evaluation}
\end{table*}

\subsection{Out-of-domain Evaluation on BEIR Benchmarks}
Following BEIR \cite{thakur2021beir}, we train retrievers with provided MS-MARCO negatives and evaluate the out-of-domain performances on 14 publicly available datasets. As is shown in Table \ref{table_beir_evaluation}, CoT-MAE v2 outperforms the previous baseline method\footnote{We do not compare with LexMAE because it did not report the results on all datasets.} by +3 points on average and achieves the best on 11 out of 14 datasets. The leading performances of CoT-MAE v2 on BEIR benchmarks show effectiveness and adaptiveness in heterogeneous retrieval task settings.

In general, comparing CoT-MAE v2 with the previous effective methods on the most commonly used benchmark datasets shows that the CoT-MAE v2 pre-training process can effectively improve dense retrieval.
The improvement derives from two aspects. \\
$\bullet$ On the one hand, CoT-MAE v2 integrates the dense and sparse representations, where dense representation naturally captures the features at the passage level, and sparse representation captures the semantics at the lexicon level, as discussed in Section \ref{Multi-view-representations}. \\
$\bullet$ On the other hand, CoT-MAE v2 uses Auto-Encoding Decoder (AE-Dec) and Auto-Regressive Decoder (AR-Dec) for the representation pre-training, where AE-Dec focuses on MLM reconstruction task and AR-dec focused on CLM generative task, providing rich self-supervision signals for the representation learning. \\
The multi-view traits of CoT-MAE v2 contribute to the efficient pre-training.

\begin{table}[!tbp]
\centering
\small

\begin{tabular}{l|c|c|c}
\toprule  
\textbf{Model} & MRR@10 & R@50 & R@1k \\
\midrule
CoT-MAE v2  & \textbf{41.4} & \textbf{89.4} & \textbf{98.7} \\

\midrule
\multicolumn{4}{c}{\textbf{Pre-traning with single-view representation bottleneck}} \\
\midrule
w/only DEN  & 40.1 (\textcolor{red}{-1.3}) & 88.0 (\textcolor{red}{-1.4}) & 98.6 (\textcolor{red}{-0.1}) \\
w/only SPR  & 40.9 (\textcolor{red}{-0.5}) & 89.0 (\textcolor{red}{-0.4}) & 98.1 (\textcolor{red}{-0.6}) \\

\midrule
\multicolumn{4}{c}{\textbf{Pre-traning with single-view decoding paradigm}} \\
\midrule
w/only AE-Dec  & 41.0 (\textcolor{red}{-0.4}) & 89.4 & 98.5 (\textcolor{red}{-0.2}) \\
w/only AR-Dec  & 40.9 (\textcolor{red}{-0.5}) & 89.2 (\textcolor{red}{-0.2}) & 98.4 (\textcolor{red}{-0.3}) \\

\midrule
\multicolumn{4}{c}{\textbf{Fine-tuning with single-view representation}} \\
\midrule
w/only DEN  & 39.4 (\textcolor{red}{-2.0}) & 87.7 (\textcolor{red}{-1.7}) & 98.7 \\
w/only SPR  & 40.2 (\textcolor{red}{-1.2}) & 88.2 (\textcolor{red}{-1.2}) & 98.4 (\textcolor{red}{-0.3}) \\

\bottomrule
\end{tabular}
\caption{Ablation study for multi-view representation and multi-view decoding paradigm pre-training. 
The scores are tested on the non-distilled retrievers.
}
\label{table_ablation_rep_dec}
\end{table}

\section{Analysis}
Given the leading performances of CoT-MAE v2, several scientific questions (\textbf{Q}) are discussed and analyzed in this section.
\paragraph{\textbf{Q1}}
How does the multi-view representation pre-training contribute to the results?

\paragraph{\textbf{Q2}}
How does the multi-view decoding paradigm contribute to the results?

\paragraph{\textbf{Q3}}
How does the performance change when using only one representation in fine-tuning and inferencing?

\paragraph{\textbf{Q4}}
What's the total dimension size of the multi-view representations? And how does performance change when shrinking the dimension of representations?
\paragraph{\textbf{Q5}}
How about the robustness of multi-view pre-training?

Corresponding answers (\textbf{A}) will be analyzed with detailed ablation data one by one as written below. 
All analyses are based on non-distilled retrievers.

\subsection{Impact of Multi-view Representations and Multi-view Decoding Paradigm}
Table \ref{table_ablation_rep_dec} shows the ablation results of multi-view representations in both the pre-training and fine-tuning stages. 

\paragraph{\textbf{A1}} As argued in Section \ref{Multi-view-representations}, multi-view representations are designed to capture sentence features from different levels, and thus achieve better performance when using them both. When only pre-training with dense or sparse representations, the performance on MS-MARCO MRR@10 drops to 40.1 and 40.9. This implies that the multi-view representation in our proposed pre-training architecture is helpful for better representation learning.

\paragraph{\textbf{A2}} As discussed in Section \ref{dec_paradigm}, the multi-view decoding paradigm provides both reconstruction and generation signals for contextual masked auto-encoding pre-training. These multi-view decoding signals promote better learning of representations. When only pre-training with AE-Dec or AR-Dec, the performance on MS-MARCO MRR@10 drops to 41.0 or 40.9. Thus the multi-view decoding paradigm promotes better representation learning.

\paragraph{\textbf{A3}} As we propose to utilize multi-view representations, both dense and sparse vectors are used in the fine-tuning and inferencing stage. Thus we need to explore how the performance changes when only using one single representation type. We try to fine-tune with only single-view representation. The performance on MS-MARCO MRR@10 drops to 39.4 and 40.2. This further implies that the multi-view representation is helpful for improving passage retrieval performance.

\begin{table}[tbp]
\centering
\small

\begin{tabular}{c|c|c|c}
\toprule  
\textbf{Dim of DEN \& SPR} & MRR@10 & R@50 & R@1k \\
\midrule
768  & 41.4 & 89.4 & 98.7 \\
384  & 40.9 & 89.1 & 98.4 \\

\bottomrule
\end{tabular}
\caption{Ablation study for reducing the dimension size of both dense and sparse vectors. 
The scores are tested on the non-distilled retrievers.
}
\label{table_ablation_dim_reduction}
\end{table}

\begin{table*}[t]
\centering
\small
\begin{tabular}{lcp{12cm}}
\toprule  
Relevant & Model  & Rank 1st passage\\
\midrule
\midrule
& & \textbf{Query:} what is green soap \textcolor{brown}{\textit{used for}} \\
 \midrule
\multirow{3}{*}{\XSolidBrush} & \multirow{3}{*}{CoT-MAE}  & \textbf{Passage:} How to Use Green Soap. Green Soap is used in many different ways. It is \textcolor{red}{\textit{made from}} pure vegetable oils and glycerine. Although not green in color, the soap is \textcolor{red}{\textit{considered green}} because it is organic, biodegradable, and environmentally safe. \\
 \midrule
 \multirow{2}{*}{\Checkmark} & \multirow{2}{*}{CoT-MAE v2}  & \textbf{Passage:} Blog - Green soap is \textcolor{blue}{\textit{used in}} tattoo and piercing studios for everything from prepping the tattoo or piercing site to cleaning tattoo machines. It's a great remover of dirt, blood, and tattoo ink. Because it's non-corrosive, it can also be \textcolor{blue}{\textit{used to}} safely soak equipment to loosen dried debris before cleaning and sterilization. \\
 \midrule
 \midrule
& & \textbf{Query:} who wrote the \textcolor{brown}{\textit{book ninjago}} \\
 \midrule
\multirow{3}{*}{\XSolidBrush} & \multirow{3}{*}{CoT-MAE}  & \textbf{Passage:} Main article: The \textcolor{red}{\textit{Lego Ninjago Movie}}. The Hageman brothers, Dan and Kevin Hageman, who wrote Lego Ninjago: Masters of Spinjitzu and co-wrote the story of The Lego Movie, will be penning the film adaptation, which will feature a new take that diverges from the TV series. \\
 \midrule
 \multirow{2}{*}{\Checkmark} & \multirow{2}{*}{CoT-MAE v2}  & \textbf{Passage:} Cole: Ninja of Earth is a \textcolor{blue}{\textit{Ninjago book}} released in 2012, \textcolor{blue}{\textit{written by Greg Farshtey}}. The book describes the adventures Cole experiences to master earth and Spinjitzu. \\
\midrule
 \midrule
& & \textbf{Query:} when was power rangers released in \textcolor{brown}{\textit{Europe}} \\
 \midrule
\multirow{3}{*}{\XSolidBrush} & \multirow{3}{*}{CoT-MAE}  & \textbf{Passage:} Main article: Originally, Power Rangers was intended to be released on July 22, 2016, a.k.a. in the latter half of the summer movie season. However, in April it was decided by the studios to  \textcolor{red}{\textit{delay}} the project until January 13, 2017 - with  \textcolor{red}{\textit{production on the movie}} scheduled to begin in January 2016.. \\
 \midrule
 \multirow{2}{*}{\Checkmark} & \multirow{2}{*}{CoT-MAE v2}  & \textbf{Passage:} Power Rangers: Super Legends has a release date of October 23rd (US) and \textcolor{blue}{\textit{2nd November in Europe}} for Wii, DS, PS2, GameCube (as the last GameCube title!) and PC. The game will celebrate Power Rangerâs 15th anniversary by gathering selectable Power Rangers from fifteen seasons of the series, from Mighty Morphin Power Rangers to Power Rangers: Operation Overdrive, and will include not only characters but Zords as well. \\
\bottomrule
\end{tabular}
\caption{Examples of rank 1st passage recalled by different models on the dev set of the MS-MARCO passage ranking dataset. Essential semantics of queries are colored in brown. We color the mismatched meanings of baseline method CoT-MAE in red, and the matching meanings of CoT-MAE v2 in blue.}
\label{table_quality}
\end{table*}

\subsection{Performances of Reducing Dimensions of Multi-view Representations}
Though excellent performance gains are achieved by CoT-MAE v2, extra dimension needs of representation are introduced due to multi-view representations. 

\paragraph{\textbf{A4}} For traditional dense retrievals methods like DPR or CoT-MAE, the dimension for one dense representation is 768. In comparison, CoT-MAE v2 further introduces a top-768 sparse vector, resulting in 2 times of representation dimension in total. Here we try to shrink down both dense and sparse vectors to 768/2=384 dimension size, which resulted in the same dimension size as DPR or CoT-MAE in total. As shown in Table \ref{table_ablation_dim_reduction}, halving the vector dimension results in a performance of MS-MARCO on MRR@10 to 40.9. But this score still far outperforms previous baselines like CoT-MAE (39.4). Thus multi-view CoT-MAE v2 shows significant robustness in the dimension-reducing scenarios.

\begin{figure}
\centering
\includegraphics[width=7cm]{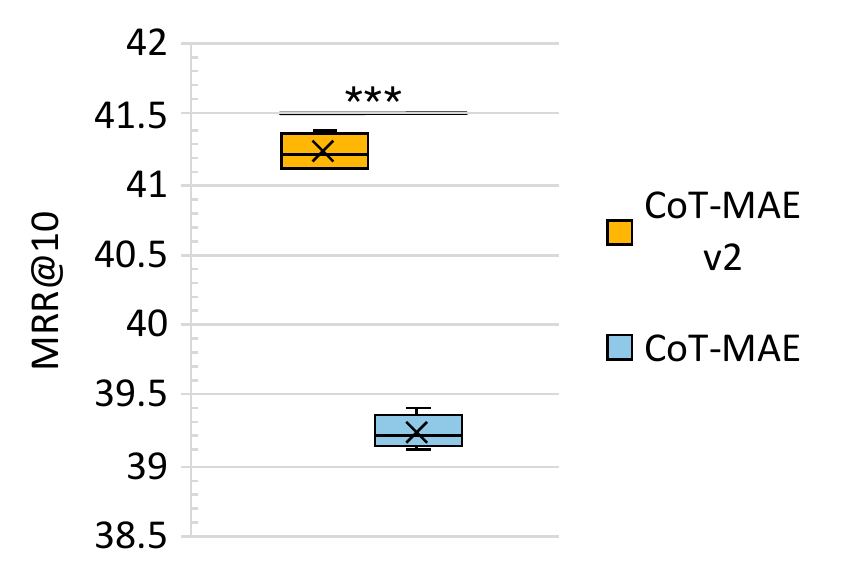}
\caption{
Comparison of multi-view pre-training performances with CoT-MAE baseline. Pre-trainings are repeated by varying random seeds. 
Scores are non-distilled retriever tested on MS-MARCO Passage Ranking Task. Student T-Test shows statistically significant results over the previous baseline.
}
\label{cotmae-v2-static}
\end{figure}

\subsection{Robustness of Multi-view Pre-training.}
\paragraph{\textbf{A5}} In order to test the robustness of our multi-view pre-training method, we repeat the pre-training for extra 3 times with different random seeds (43, 2022, 2023). As is shown in Figure \ref{cotmae-v2-static}, CoT-MAE v2 clearly outperforms its previous baseline with good performance stability, showing its robustness and effectiveness.

\subsection{Case Study}

To qualitatively understand the gains from our proposed pre-training method, we pick up some examples from the first recalled passages of CoT-MAE v2, comparing them with its baseline method CoT-MAE in Table \ref{table_quality}. \\
$\bullet$ The first query is related to the usage ``\textit{used for}'' (marked in brown). CoT-MAE baseline misses the relevant answers but returns a passage that tells contents about ``\textit{made from}'' and ``\textit{considered green}'' (marked in red). In comparison, CoT-MAE v2 retrieves the accurate answers (marked in blue). \\
$\bullet$ The second query asks the writer of ``\textit{book}'' (marked in brown). CoT-MAE baseline returns the passage about ``\textit{movie}'' (marked in red). In comparison, CoT-MAE v2 retrieves the ``\textit{Ninjago book}'' with its author name (marked in blue). \\
$\bullet$ The third query asks for the ``\textit{release date of power rangers in Europe}'' (marked in brown). Although the CoT-MAE baseline retrieves a passage related to the power rangers, this passage only tells the ``\textit{original release date, delay schedule, and movie production date}'' (marked in red). None of them tells the answers to the release date in Europe. In comparison, our CoT-MAE v2 exactly retrieves the real answers ``\textit{2nd November in Europe}'' (marked in blue).

These examples show that CoT-MAE v2 can capture more precise semantic meanings in passages, demonstrating that the CoT-MAE v2 pre-training method is more effective than the previous pre-training methods.

\section{Conclusion}
This paper proposes a multi-view contextual masked auto-encoding pre-training architecture for better passage retrieval. Experiment results show that multi-view representation and multi-view decoding paradigms significantly contribute to effective retrieval performance. Our method also shows good robustness and stability. In the future, we will further explore incorporating new pre-training paradigms to get more effective and robust retrievers.

\newpage

\bibliography{anthology,custom}
\bibliographystyle{acl_natbib}

\appendix
\end{document}